\documentclass[11pt]{article}

\usepackage[preprint]{acl}

\usepackage{times}
\usepackage{latexsym}
\usepackage{url}
\usepackage{comment}
\usepackage[T1]{fontenc}
\usepackage[table]{xcolor}
\usepackage[utf8]{inputenc}

\usepackage{microtype}
\usepackage{amsmath}
\usepackage{todonotes}

\usepackage{inconsolata}

\usepackage{graphicx}

\title{\textsc{LLMberjack}: Guided Trimming of Debate Trees \\for Multi-Party Conversation Creation}

\author{
  \textbf{Leonardo Bottona\textsuperscript{1,2}},
  \textbf{Nicol\`o Penzo\textsuperscript{1,2}},
  \textbf{Bruno Lepri\textsuperscript{2}},
  \textbf{Marco Guerini\textsuperscript{2}},
  \textbf{Sara Tonelli\textsuperscript{2}}
\\
  \textsuperscript{1}University of Trento
  \textsuperscript{2}Fondazione Bruno Kessler
\\
  \small{
    \textbf{Correspondence:} \href{mailto:email@domain}{leonardo.bottona@studenti.unitn.it}, \href{mailto:email@domain}{npenzo@fbk.eu}
  }
}

\begin{document}
\maketitle

\begin{abstract}
Online debates on platforms like Reddit or Kialo are structured as \textit{reply trees}: each message can receive several replies, which in turn can each receive several replies, producing a branching structure rather than a linear conversation. Turning one path (or a mix of paths) through such a tree into a natural, turn-by-turn conversation among multiple speakers (a process we call \textit{linearization}) requires deciding which messages to keep, in what order, and who each message is addressed to. To this aim, we present \textsc{LLMberjack}, a platform for creating multi-party conversations starting from existing debates, originally structured as reply trees. The system offers an interactive interface that visualizes discussion trees and enables users to construct coherent linearized dialogue sequences while preserving participant identity and discourse relations. It integrates optional large language model (LLM) assistance to support automatic editing of the messages and speakers' descriptions. We provide preliminary evidence about the platform’s utility by showing how tree visualization facilitates the creation of coherent, meaningful conversation threads and how LLM support enhances output quality while reducing human effort. The tool is open-source and designed to promote transparent and  reproducible workflows to create multi-party conversations, addressing a lack of resources of this type.
\end{abstract}

\section{Introduction}\label{sec:intro}

Despite ongoing efforts in the NLP community to create large datasets and linguistic resources, there is traditionally a lack of high-quality datasets with multi-party conversations, shortened as MPCs \cite{penzo-etal-2024-llms}.
Platforms such as X, Reddit and Kialo provide a large amount of conversations in the form of \textit{reply trees}: a root message can receive several replies, each of those replies can receive further replies, and so on, producing a branching structure rather than a single back-and-forth exchange. 
\begin{figure}
    \centering
    \includegraphics[width=1.0\linewidth]{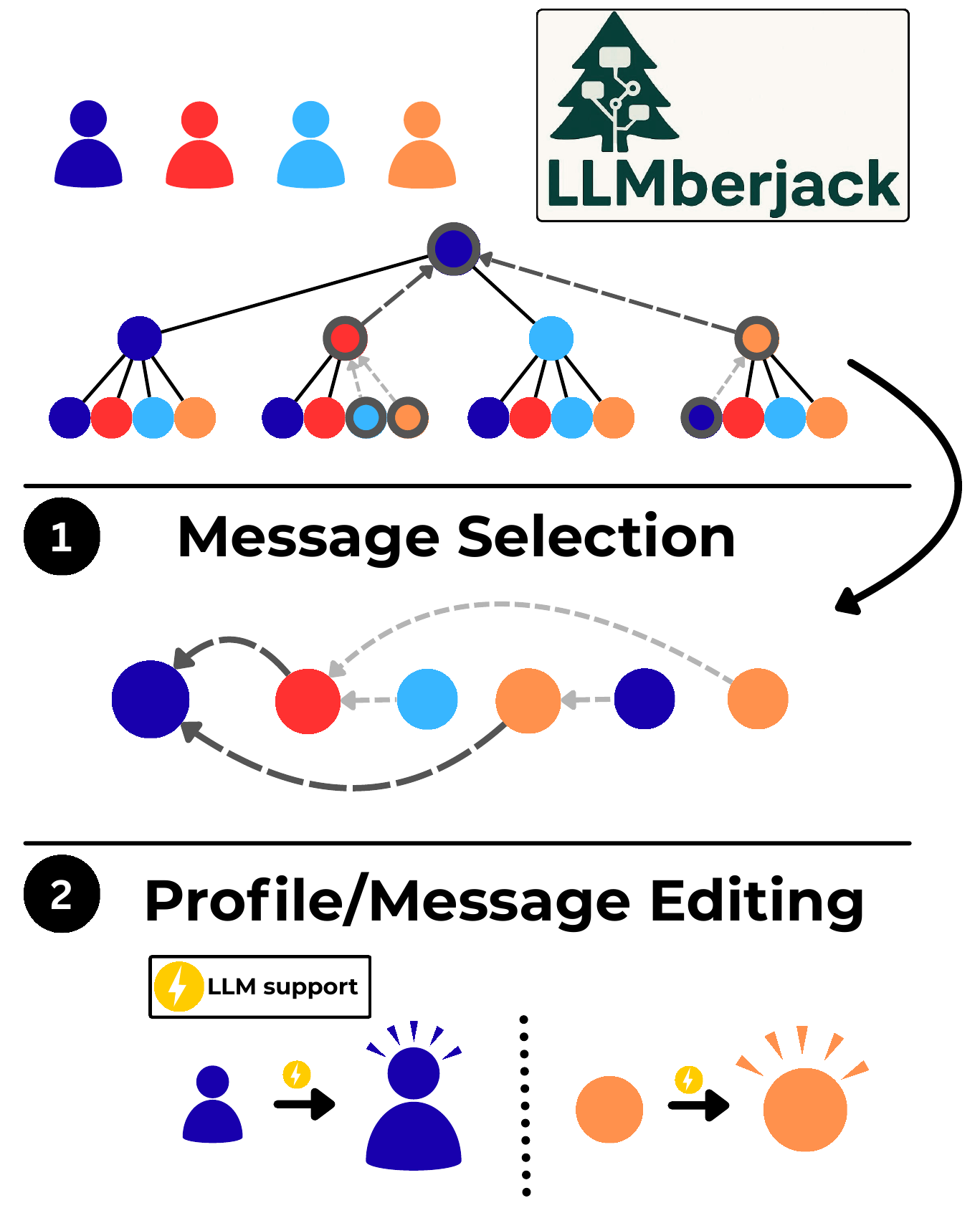}
    \caption{Overview of the \textsc{LLMberjack} platform. The interface integrates reply-tree visualization, message selection tools for building linearized multi-party conversations (1), and LLM-support for editing messages and speaker profiles (2).}
    \label{fig:first_image}
\end{figure}
For example, a root post ``Should we introduce a four-day work week?'' might receive two replies, one from a user arguing in favor and one arguing against, and each of these replies might in turn receive its own replies, so that the discussion develops along multiple, sometimes independent, branches. Each root-to-leaf path in such a tree can be read top to bottom as a multi-party conversation \cite{derczynski-etal-2017-semeval, penzo-etal-2024-putting}, since in this way it clearly shows who wrote each message and which earlier message it replies to; however, a single path rarely captures messages that address more than one participant at a time. We call the process of selecting and ordering messages from a reply tree into such a turn-by-turn conversation \textit{linearization}.

Messaging platforms like Telegram and WhatsApp, instead, present inherently linear conversations that often contain overlapping or parallel sub-dialogues, frequently with multiple implicit addressees for each message. So, while representing examples of MPCs, an annotation step would still be needed to make addressees (i.e., who a message is directed to) explicit and enable modelling their complex conversation structures. Furthermore, using discussions from online platforms to study MPCs raises significant privacy and profiling concerns \cite{10.5555/3666122.3667033}.

LLMs could be potentially used to address the lack of MPC datasets by generating synthetic dialogues. 
However, as shown by \citet{penzo2025dontstopmultipartygenerating}, although some LLMs can produce high-quality synthetic dialogues, they may still struggle with the generation of complex structures with multiple speakers.

A possible solution to create linearized multi-party conversations with overlapping or parallel sub-dialogues starting from existing reply trees is to have human annotators traverse the tree following the reply-to links between messages, selecting at each step which message to include next and deciding who it is addressed to. 
Annotators can also edit message content or redefine addressee links along the way, thereby enhancing both naturalness and interactional coherence. Furthermore, a single reply tree can yield several linearized MPCs, capturing potential conversation variations that result from different turn-taking orders.

In this paper, we introduce \textsc{LLMberjack}, a Human-AI collaborative platform designed to create synthetic, thread-like multi-party conversations starting from existing reply trees. The platform provides an interface that allows annotators to traverse the tree step by step, visualizing both the parent and child nodes for each message, thereby making selection decisions more context-aware.

Reply trees extracted from structured debate platforms like Kialo\footnote{\url{https://www.kialo.com/}} or automatically generated may exhibit a style that is not fluent or natural enough. To enhance specific linguistic features or user traits, we implement an LLM-assisted protocol that supports two key tasks besides tree editing: \textsc{(i.)} \textit{speaker profiling}, i.e., generating a short description of a speaker's conversational style and personality based on their messages (either from the emerging conversation or from the original reply tree), and merging it with any pre-existing description of that speaker; \textsc{(ii.)} \textit{message editing}, i.e., the LLM refines a given message by considering the chat history and speaker profile. Human annotators then decide whether to accept, modify or reject the LLM's suggestion, ensuring the overall conversational quality.

We evaluate the impact of both tree visualization and LLM-assisted message editing involving four annotators.  Results suggest that the quality of the resulting MPCs improves when tree visualization is available, and that LLMs  can effectively support message editing, while also accelerating the annotation process.

\textsc{LLMberjack} is available on a dedicated GitHub repository\footnote{\url{https://github.com/LeonardoBottonaUniTn/llmberjack}}.
The platform targets researchers from NLP and Social Sciences, helping them in the creation of high-quality MPCs with specific characteristics.

\section{Related Work}

Multi-party conversational corpora have been collected from a broad range of environments, including in-person meetings \cite{10.1007/11677482_3, 1198793} and online platforms \cite{ouchi-tsuboi-2016-addressee, zhang-etal-2018-conversations, chang-danescu-niculescu-mizil-2019-trouble}. However, these diverse sources exhibit different characteristics that complicate cross-domain generalization. 
For instance, spoken multi-party dialogues are heavily shaped by non-verbal cues, physical setting, and overlapping turns, all of which are typically absent in written online interactions. Conversely, text-based conversations unfold asynchronously, without overlap, and often follow platform-specific conventions that further influence interaction patterns \cite{mahajan-shaikh-2021-need, penzo2025dontstopmultipartygenerating}. Heterogeneity in structure and annotation practices is shown also across datasets from similar sources.

The limited availability of multi-party conversation data with the desired level of structural and interactional detail suggests the need for alternative approaches. One promising direction is the use of \emph{synthetic}, human-in-the-loop methods, which allow researchers to control conversational conditions while preserving interactional plausibility. This has been tested in single-turn interactions \cite{fanton-etal-2021-human, russo-etal-2023-countering} and for multi-turn dialogues \cite{bonaldi-etal-2022-human, occhipinti-etal-2024-fine}, but not yet for multi-party settings. Only \citet{chen-etal-2023-places} and \citet{penzo2025dontstopmultipartygenerating} have attempted to generate synthetic multi-party conversations, the former involving up to three users and the latter extending to interactions among four to six users. 

\citet{menini-etal-2025-first} introduced \textsc{FirstAID}, a platform for the synthetic creation of document-grounded dialogues among multiple participants, whose evaluation however has been limited to 1-to-1 interactions. Among the existing tools, \textsc{ConvoKit} \citep{chang-etal-2020-convokit} is the most established one for multi-party settings, which focuses on the analysis of conversations. 
Yet, despite these contributions, there is still no open-source platform that supports the \emph{creation} with \emph{human-AI refinement} of multi-party conversations from structured reply trees. 
\textsc{LLMberjack} is, to the best of our knowledge, the first open-source platform explicitly designed for this task, combining tree-based message selection with LLM-assisted speaker profiling and message refinement.

\begin{figure*}
    \centering
    \includegraphics[width=0.98\linewidth]{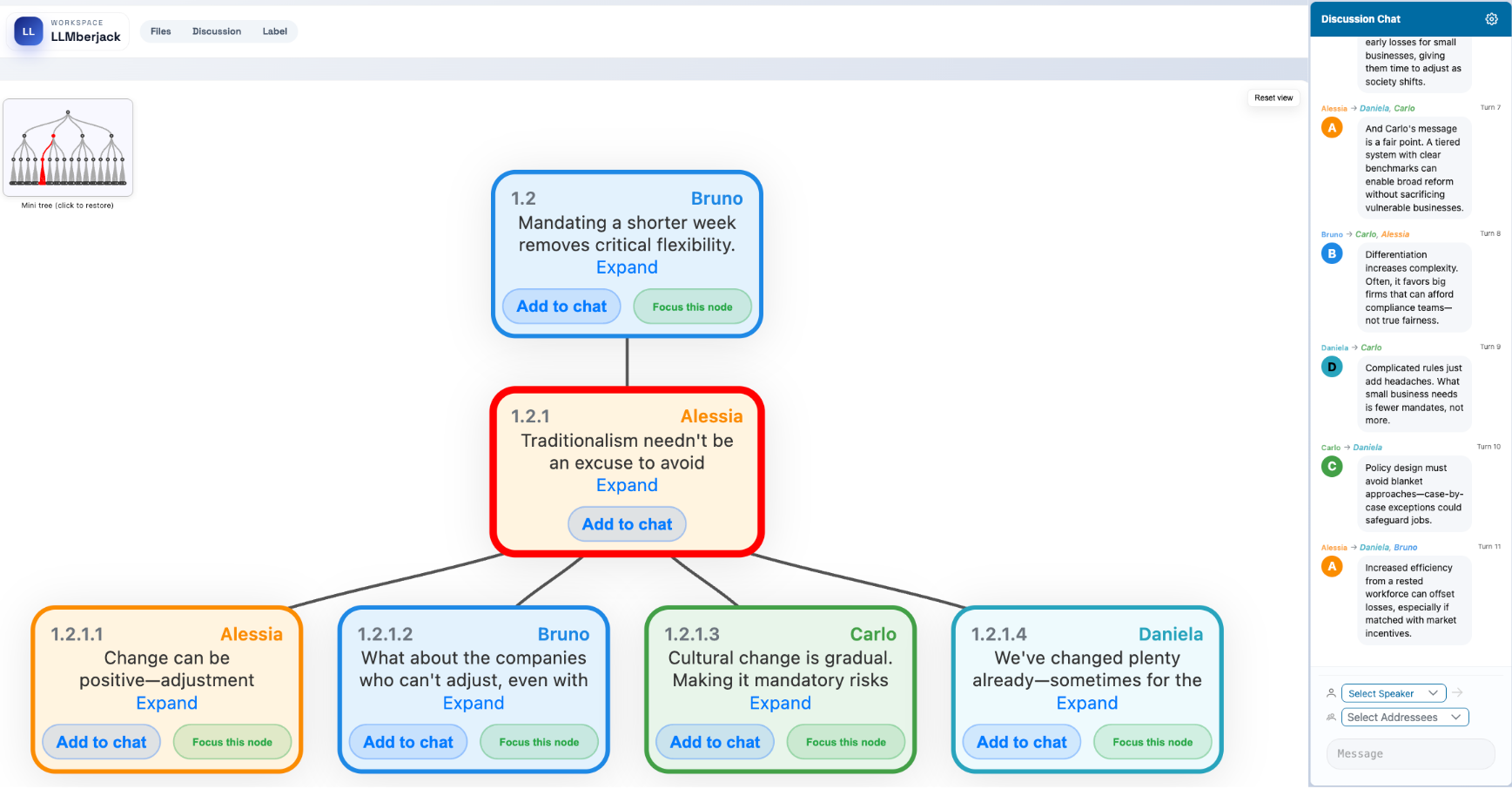}
    \caption{Screenshot of tree visualization for node 1.2.1 (left) and of the chat creation tab (right). Each node-box reports the speaker's name on the top-right corner, and a preview of the message in the center (expandable).}
    \label{fig:screen_tree}
\end{figure*}

\section{System Architecture}

\textsc{LLMberjack} is designed to support the full process of  transforming structured reply trees into coherent multi-party conversations. The system is organized into three main layers: \textsc{(i.)} a data-processing backend, \textsc{(ii.)} an interactive data manipulation interface, and \textsc{(iii.)} an export module.

\subsection{Data Processing Backend}

\par{\textbf{Tree-Centric Data Model.}}
All discussion sources are represented as rooted reply trees. Each node corresponds to an individual message and stores author and text of message, and other existing  platform-specific attributes, if any. Edges encode explicit reply-to relations.
\par{\textbf{Backend Services.}}
The backend provides: \textsc{(i.)} parsing routines that convert raw \texttt{json} dumps into the internal graph representation; \textsc{(ii.)} subtree querying for efficient visualization and traversal; \textsc{(iii.)} 
file-management functionalities for uploading discussion files, performing LLM-assisted tree normalization when the structure is imperfect, and handling draft files containing partial or previously linearized conversations; \textsc{(iv.)} LLM endpoints for message refinement and speaker profiling. 

\subsection{Interactive Data Manipulation Interface}

The data manipulation environment is implemented using \textit{Vue.js} and \textit{D3.js} to provide real-time synchronization between the debate tree and the emerging linearized conversation.

\par{\textbf{Tree Visualization.}}
Annotators are presented with an interactive view of the full debate tree featuring: \textsc{(i.)} a global structural visualization of the entire debate tree and a focused node view showing the selected node together with its parent and children; \textsc{(ii.)} color-coded authors.
We report a screenshot of the visualization in Figure \ref{fig:screen_tree}. This view facilitates the exploration of alternative conversational paths and supports informed linearization decisions.

\par{\textbf{Thread Construction.}}
Annotators construct linear sequences of conversation turns by selecting messages from a given reply tree and placing them in a turn-by-turn order. The interface allows annotators to reorder messages, redefine addressee relations (for example by selecting multiple addressees for a turn) and enforce soft constraints (e.g., minimal edits, conversational plausibility). 

\par{\textbf{Labelling Support.}} The platform additionally provides a specific window for message or conversation-level labelling, supporting free-text labels and binary or Likert scale ratings, intended to support future evaluation workflows of this kind (details in Appendix \ref{app:annot_interf}).

\subsection{LLM-Assisted Refinement Module}

\begin{figure}
    \centering
    \includegraphics[width=0.98\linewidth]{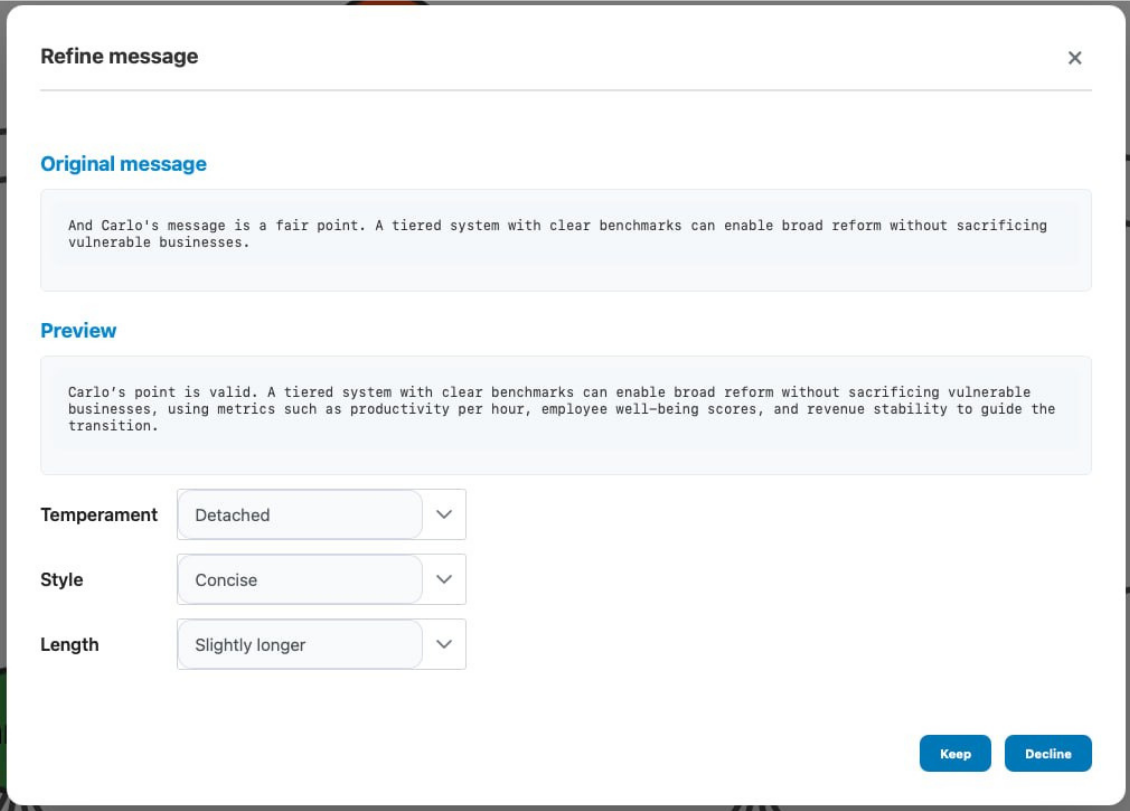}
    \caption{Screenshot of the LLM-assisted message refinement page.}
    \label{fig:screen_refine}
\end{figure}

\paragraph{Speaker Profiling.}
Each user is associated with a speaker profile, either provided as input or assigned as a default version when unavailable (details in Appendix \ref{app:data_management}).
Upon request, the platform refines such profiles using an LLM. We use \texttt{Llama 4 Maverick} \cite{meta2025llama} for all the LLM-assisted tasks, exploiting the \texttt{Groq} cloud platform\footnote{\url{https://groq.com/}}. We select \texttt{Llama 4 Maverick} since it is one of the most widely-used instruction-following LLMs and could be easily run on the Groq cloud platform for free.\footnote{The experiments have been performed before the deprecation of this model from Groq (\url{https://console.groq.com/docs/deprecations})} However, our platform is model-agnostic by design, and users can configure their own API credentials and choose among multiple supported LLM providers and models (prompts, generation and implementation details in Appendix \ref{app:llm_integr}).

To construct or refine a profile, the model receives: \textsc{(i.)} the original speaker profile to be refined, and \textsc{(ii.)} a set of selected messages from the speakers serving as contextual evidence (details in Appendix \ref{app:llm_integr}). Based on this information, the LLM infers stylistic patterns and conversational temperament merging them into an updated profile.

\par{\textbf{Message Refinement.}}
The LLM can refine individual messages under annotator's control. The model receives: \textsc{(i.)} the message to refine; \textsc{(ii.)} the local conversational context, i.e., all messages appearing before the one being refined; \textsc{(iii.)} the speaker profile; \textsc{(iv.)} the constraints set by the annotation protocol (style, length, temperament). Based on this information, the LLM generates a new, improved version of the original message.

The annotator can accept or modify the proposed revision, ensuring the final version remains coherent and free from hallucinations or stylistic drift. We report a screenshot of the message refinement page in Figure \ref{fig:screen_refine}.

\subsection{Data Export, Deployment and Availability}
The system currently provides \texttt{json} export for the final conversations.
The full platform is publicly available as open-source software via the dedicated GitHub repository mentioned in Section \ref{sec:intro}. It can be deployed \emph{locally}, for secure or small-scale dataset creation tasks. 
The repository includes installation scripts, configuration templates, and a demo instance, facilitating adaptation to diverse annotation protocols and datasets. 

\section{Evaluation of \textsc{LLMberjack} Features}

We evaluate through human assessment the impact of two core features of \textsc{LLMberjack}: the impact of tree visualization on the creation of MPCs and the LLM-assisted message editing. To isolate their effects, we split the analysis into two parts. Firstly, we perform the message selection task, starting from a reply tree, comparing conditions with and without tree visualization. Secondly, we edit a subset of messages, comparing scenarios with and without LLM support. 
The evaluation is conducted at a modest scale, under a counterbalanced design intended to reduce annotator- and topic-specific confounds (see Limitations).

\subsection{Creation of Synthetic Reply-Trees}\label{subsec:synt_reply_trees}
 
As a preliminary step for our evaluation, we first generate synthetic reply-trees. 

Specifically, for each synthetic reply-tree we first ask the LLM to define a set of $4$ users and then  generate iteratively the full tree. The process starts with one single initial message from a random user (i.e., the root of the discussion), followed by one reply from each participant (including the self-replies). This procedure is repeated recursively for each new node up to a specified depth ($4$ in our case), resulting in a total of $85$ messages. LLMs are generally proficient at producing coherent one-to-one replies that respect speaker profiles or conversational roles. From these generated reply trees, we make the annotators build linearized multi-party conversations.

We generate $4$ synthetic reply trees using GPT-4.1\footnote{\href{https://platform.openai.com/docs/models/gpt-4.1}{platform.openai.com/docs/models/gpt-4.1}}. Each reply tree is about a different topic. For each topic, two speakers were assigned a pro stance and the other two speakers a counter stance with respect to the topic. We report the selected topics and further details in Appendix \ref{app:reply_trees}.

The evaluation process consists of two main steps: \textsc{(i.)} selection of messages to build a MPC starting from the synthetic reply tree, with and without tree visualization (Section \ref{subsec:expla_firststep}); \textsc{(ii.)} editing of the resulting MPC messages, with and without LLM support (Section \ref{subsec:expla_secondstep}).

\subsection{Evaluating the Impact of Tree Visualization}\label{subsec:expla_firststep}

Annotators were asked to create a multi-party conversation from a given synthetic reply tree by selecting a subset of nodes/messages. They were instructed to follow the rules below.

\par{\textbf{R\textsubscript{1}:}} 
The opening message must be a general statement on the given topic addressed to everyone.
\par{\textbf{R\textsubscript{2}:}} Each conversation should contain between 10 and 15 messages and should resemble the style of a typical Telegram chat.
\par{\textbf{R\textsubscript{3}:}} Annotators may change or add addressee relations at their discretion but all users must contribute at least one turn. The tree structure serves as a suggestion rather than a strict constraint.
\par{\textbf{R\textsubscript{4}:}} Annotators should perform only minimal, necessary edits to correct errors or ensure conversational flow. Messages should not be edited to improve style or argumentative quality, which will be part of the second evaluation step (Section \ref{subsec:expla_secondstep}).

Annotators were asked to create $3$ distinct MPCs from each tree, aiming for variation in content and interaction patterns across conversations. 
Before starting the main task, each annotator was instructed to read all speakers' profiles carefully.
Annotators completed the task under two different visualization conditions: option \textit{w Tree}, which provided full access to the tree-structure visualization during MPC creation, and option \textit{w/o Tree}, which presented all the messages as a single flat sequence without tree visualization. 
For each of the four synthetic reply trees, two annotators performed the task \textit{w Tree} visualization and two \textit{w/o Tree} visualization. We report further details of the annotation process in Appendix \ref{app:step1}.

After the MPCs were created, two independent evaluators assessed their quality through pairwise comparisons of sets of conversations produced from the same reply tree, created \textit{w Tree} or \textit{w/o Tree} visualization, for a total of $16$ pairs. Each comparison was performed along three dimensions:

\par{\textbf{1. Naturalness of the conversation}}, focusing on the coherence of the conversational flow, the plausibility of turn-by-turn progression, and the overall smoothness of the dialogue.
\par{\textbf{2. Conversation Variability}}, assessing whether the set of conversations derived from the same tree exhibited meaningful diversity in content, interaction patterns, and turn-taking structure.
\par{\textbf{3. Participants' Engagement}}, evaluating the degree to which the conversation goes beyond generic statements and displays targeted, socially meaningful exchanges. This includes the presence of distinctive interactional behaviors, speaker-specific styles, responsive turns that directly engage with previous messages, and interactional patterns that make the dialogue feel lively, purposeful, and contextually grounded.

For each dimension, evaluators indicated which conversation in the given pair they considered of higher-quality or whether the two were equivalent.

\paragraph{Quantitative Evaluation.} In Table \ref{tab:quality_trees}, we report evaluation results for the $3$ dimensions above, the average turn-selection speed in terms of turns/minute (v\textsubscript{turn}) and the inter-annotator agreement using Weighted Cohen's kappa ($\kappa_{w}$). The results show an advantage for the \emph{w Tree} condition over the \emph{w/o Tree} setting (only for the Variability dimension there is a relative majority of ties). This advantage is particularly evident for the Naturalness dimension. Furthermore, the average speed increases by almost $25\%$ \emph{w Tree} visualization. Agreement ranges from $0.25$ to $0.44$, highlighting the subjectivity of the annotation task. 

\begin{table}[]
    \centering
    \small
    \begin{tabular}{l | c c c | c }
    \hline
         & \textbf{Nat.} & \textbf{Var.} & \textbf{Eng.} & \textbf{v\textsubscript{turn}} \\
        \hline
        \hline
        \textbf{w Tree} & \textit{\small{65.62}} & \small{34.37} & \textit{\small{49.99}} & \textit{\small{1.82}}\\
        \hline
        \textbf{w/o Tree} & \small{28.13} & \small{21.88} & \small{28.13} & \small{1.46}\\
        \hline 
        \textbf{\textit{tie}} & \small{6.25} & \textit{43.75} & \small{21.88} & \small{/}\\
        \hline
        \hline
        $\kappa_w$ & 0.44 & 0.40 & 0.25 &  / \\
        \hline
    \end{tabular}
    \caption{Percentage of MPC comparisons where one setting (with or without tree visualization) was preferred over the others in terms of naturalness (Nat.), variability (Var.), and participants' engagement (Eng.). The last column reports the average turn-selection speed in turns/minute ($v_{\text{turn}}$). The final row shows inter-annotator agreement (weighted Cohen’s $\kappa_{w}$).}
    \label{tab:quality_trees}
\end{table}

\paragraph{Qualitative Observations.} We also collected all the feedback and comments provided by the evaluators during the sessions. They reported that conversations created with tree visualization tended to focus on fewer subtopics but developed them more deeply, exhibiting richer argumentative structure and stronger relational coherence across messages. On the contrary, conversations produced without tree visualization typically covered a broader range of aspects of the main topic but remained more superficial in their argumentative depth. In general, they confirmed the difficulty in identifying a version of higher quality than the other, since quality was generally high among all the given conversations. Annotators consistently reported that the tree visualization was substantially more helpful for the task. They appreciated the implicit “guidance’’ it provided, allowing them to make more confident and reliable choices, particularly about choosing the addressee(s). Annotators noted that the visualization would be even more advantageous in larger annotation rounds (more than $3$), as it facilitates the identification of multiple plausible MPCs through different traverses from the same debate tree and reduces cognitive effort during the task.

\paragraph{Structural Metrics.} Table~\ref{tab:structural-metrics} summarizes structural properties of the created MPCs, comparing the \textit{w Tree} and \textit{w/o Tree} conditions. The number of messages per MPC and the number of addressees per turn are comparable across the two conditions, indicating that tree visualization does not change how much a conversation grows or how broadly turns are addressed. In contrast, the interleaving score, which measures how often consecutive turns switch between different branches of the original reply tree\footnote{We measured the interleaving score as the fraction of consecutive turn-pairs, taken in chronological order, whose belong to different first-level branches.}, is higher \textit{w/o Tree} ($0.37$ vs. $0.22$), showing that annotators switch between first-reply branches more often when the tree structure is not visible, while \textit{w Tree}, MPCs tend to stay within a single branch for longer before switching. 

\begin{table}
        \centering
        \small
    \begin{tabular}{l|c|c}
    \hline
        \textbf{Metric} & \textbf{w Tree} & \textbf{w/o Tree} \\
         \hline
          \hline
        \textbf{Avg Number of Turns}                    & $12.17^{\pm 1.43}$ & $12.42^{\pm 1.47}$ \\
        \hline
        \textbf{Avg Addressees per Turn}         & $1.53^{\pm 0.72}$  & $1.57^{\pm 0.73}$  \\
        \hline
        \textbf{Avg Interleaving score}          & $0.22^{\pm 0.21}$ & $0.37^{\pm 0.26}$ \\
        \hline
    \end{tabular}
    \caption{Structural metrics of the MPCs created with and without tree visualization (mean $\pm$ standard deviation). Number of turns per MPC and addressees per turn describe conversation size and engagement; the interleaving score describes how often turn order switches between first-level branches of the original reply tree, with higher values indicating more frequent switching.}
    \label{tab:structural-metrics}
\end{table}

\subsection{Evaluating the Impact of LLM Support}\label{subsec:expla_secondstep}
In the second evaluation step, we aim to assess the effect of LLM-assisted message editing compared to the editing without LLM support. 
$4$ annotators refined a total of $8$ conversations (two conversations for each topic). For each annotator--topic combination, one conversation was edited  with LLM assistance and the other without. All four annotators worked on every conversation, and for each conversation, two used LLM support while the other two performed the task manually. 

To ensure a controlled experimental setup and avoid fully rewriting the given conversations, each annotator was instructed to focus only on one speaker and to edit, if needed, only his/her messages throughout a given conversation. 
The editing should specifically involve \emph{style}, \emph{temperament}, and \emph{length}.

After the MPCs were edited, two evaluators assessed their quality by comparing, for the same MPC, the conversations edited  \emph{w LLM} assistance against the versions refined without it (\emph{w/o LLM}), for a total of $32$ pairs. Each pair of conversations was evaluated along two dimensions:

\par{\textbf{1. General turn quality}}, considering both the coherence of each turn 
    and its contribution to the conversation flow;
\par{\textbf{2. Adherence to the refinement requirements}}, evaluated across the three 
    specified sub-dimensions: \emph{length}, \emph{temperament}, and \emph{style}.

For each dimension, annotators indicated whether the \emph{w LLM} support or \emph{w/o LLM} editing was of better quality, or whether the two versions were considered equivalent.
Details about task and evaluation are reported in Appendix \ref{app:step2}.

\paragraph{Quantitative Evaluation.} In Table \ref{tab:quality_refinement}, we report the evaluation results together with the average refinement velocity in terms of tokens\footnote{Number of tokens of the final refined sentence}/second (v\textsubscript{tokens}). Overall, the results show an advantage for the \emph{w LLM} condition compared to the \emph{w/o LLM} setting. 
 The average refinement velocity indicates that LLM support speeds up the refinement process by approximately $83\%$. 
 Agreement ranges from $0.36$ to $0.58$, highlighting also here the subjectivity of the annotation task, except for the Length dimension (which is intuitively more objective).

\paragraph{Qualitative Observations.} Feedback from the evaluators confirmed that annotators' experience with linguistic tasks has an important impact on the quality of refinements, regardless of whether LLM assistance is provided, 
particularly for dimensions such as \emph{style} and \emph{temperament}. Nonetheless, they consistently noted that LLM support is crucial when generating substantially longer messages, where manual refinement alone is often more challenging. Annotators agreed that the LLM is particularly helpful for reorganizing sentences rather than making minor additions or deletions, a crucial aspect for longer messages. At the same time, they noted that the LLM occasionally introduces repetitive interjections; still, with minimal human editing, these issues can be easily fixed.

\begin{table}[]
    \centering
    \small
    \begin{tabular}{l | c | c c c | c}
    \hline
         & \textbf{Gen.} & \small{\textbf{Len.}} & \small{\textbf{Style}} & \small{\textbf{Temp.}}  & \textbf{v\textsubscript{tokens}}\\
        \hline
        \hline
        \textbf{w LLM} & \textit{64.06} & \textit{57.81} & \textit{64.06} & \textit{56.25} & \textit{0.86}\\
        \hline
        \textbf{w/o LLM} & \small{17.19} & \small{4.69} & \small{25.00} & \small{31.25}  & 0.47\\
        \hline
        \textbf{\textit{tie}} & \small{18.75} & \small{37.50} & \small{10.94} & 12.50 & \small{/}\\

        \hline
        \hline
        $\kappa_w$ & 0.36 & 0.58 & 0.43 & 0.44  & / \\
        \hline
    \end{tabular}
    \caption{Percentage of times one setting (with or without LLM support) was preferred over the other in terms of general quality (Gen.), length (Len.), style (Style), and temperament (Temp.), together with the average refinement speed in tokens/second ($v_{\text{tokens}}$). The final row reports inter-annotator agreement (weighted Cohen’s $\kappa_{w}$).}
    \label{tab:quality_refinement}
\end{table}

\section{Conclusion}
In this paper, we presented \textsc{LLMberjack}, a Human-AI collaborative platform designed to generate synthetic thread-like multi-party conversations starting from tree-structured debates, with optional LLM support for message refinement. Our goal is to alleviate the scarcity of high-quality MPC datasets with well-controlled interactional and structural properties by providing annotators with an intuitive interface that supports more guided and more consistent decision-making. 
Within this setting, our evaluations provide preliminary evidence that the platform can accelerate the overall creation workflow, both in message selection and in refinement, while also leading to conversations of higher or equal quality.

\section*{Limitations}
Our evaluation is conducted at a modest but tightly controlled scale: four synthetic reply trees, four annotators, and two evaluators, yielding 16 pairwise comparisons for the tree-visualization condition (Section \ref{subsec:expla_firststep}) and 32 for the LLM-support condition (Section \ref{subsec:expla_secondstep}). The counterbalanced design reduces annotator- and topic-specific confounds, but a sample of this size is not intended to support strong statistical claims: we do not report significance tests or confidence intervals, and the percentages in Tables \ref{tab:quality_trees} and \ref{tab:quality_refinement} should be read as descriptive summaries of this specific sample. We view these results as promising, preliminary evidence, with replication across more reply trees, topics, and annotators as a natural next step to establish which differences are most robust.

Moreover, our evaluation setup involves some specific choices whose generalizability could be further tested. The reply trees were synthetically generated with a single LLM (GPT-4.1) on topics salient to an italian audience, and annotators and evaluators were italian speakers based at the same institutions, which allowed for a controlled, well-understood setting but leaves open how results transfer to other languages, topics, or annotator pools. Similarly, the LLM-assisted platform feature for message refinement was evaluated using a single model (\texttt{Llama 4 Maverick}); the platform itself is model-agnostic by design, and testing additional providers and model sizes is a valuable direction for future work.

Finally, the current evaluation focuses specifically on isolating the effect of tree visualization and of LLM support within \textsc{LLMberjack}, rather than on benchmarking the platform as a whole. As such, it does not yet include reply trees from real online debates, which could introduce additional challenges such as uneven reply depth or off-topic branches, nor a comparison against external baselines or related tools such as \textsc{ConvoKit} \citep{chang-etal-2020-convokit} or \textsc{FirstAID} \citep{menini-etal-2025-first}. Evaluating the platform on real debate data and against alternative workflows is a promising direction for follow-up work.

\section*{Ethics Statement}
All annotators and evaluators involved in the data collection/evaluation were hired as PhD students or postdoctoral researchers at one of the institutions involved. The synthetic reply trees given to the annotators were carefully analyzed at the beginning to check the possible presence of offensive language or toxic content. No personal data were used to conduct this study and the speaker profiles were fully synthetic. Still, we are aware of the potential data leakage from LLM training data. This platform can help to paraphrase also real conversations for pseudo-anonymization purposes.

\section*{Acknowledgements}

We thank Genoveffa Martone, Daniela Occhipinti, Daniel Russo and Sebastiano Vecellio Salto for their contribution to the evaluation activities. We thank Helena Bonaldi, Stefano Menini and Daniel Russo for the suggestions. The work of Bruno Lepri was partially supported by the NextGenerationEU Horizon Europe Programme, grant No. 101120237 -- ELIAS and grant number 101120763 -- TANGO. The work of Marco Guerini was partially supported by the European Union’s CERV fund under grant agreement No. 101143249 -- HATEDEMICS.
Nicolò Penzo’s activities have been part of the network of excellence of the European Laboratory for Learning and Intelligent Systems (ELLIS).  Leonardo Bottona's work was developed during his internship in Fondazione Bruno Kessler (Italy). The authors also thank AitomotiveLab for supporting Leonardo Bottona's work.

\bibliography{custom}

\clearpage

\appendix

\section{Technical details}\label{sec:appendix}

\subsection{System Architecture}
\textsc{LLMberjack} adopts a client--server design.
The front-end (Vue.js + D3.js) handles visualization and user interaction,
while a Python backend manages data structures, annotation logic, 
and controlled LLM calls. Components communicate through a RESTful API.

\subsection{Data and File Management}\label{app:data_management}
Discussion files are represented as rooted trees whose nodes store message text, author metadata, and parent/child links. The system supports two file types: \textit{discussion files} (full debate trees) and \textit{draft files}
(partially or fully linearized conversations).  
If a discussion file has an imperfect or noisy structure, users may invoke an
LLM-assisted normalization step that reconstructs missing or inconsistent reply
relations. When the \texttt{users} section is missing or incomplete, the system
automatically extracts all speakers from the debate tree and regenerates the
\texttt{users} list, assigning each participant a default profile with the
description ``This is a telegram user''.

\subsection{LLM Integration}\label{app:llm_integr}
LLM calls follow fixed templates. For speaker profiling, the model receives the speaker profile to refine and a set of selected messages from the speakers serving as contextual evidence. Such contextual evidence corresponds either to the speaker's
messages from the emerging linearized conversation (if at least three messages from the speaker are written) 
or all nodes authored by that speaker in the original reply tree.  
For message refinement, the LLM is given the message to edit, the speaker profile,
and the local conversational context, i.e., all turns preceding the one being refined.
 
For \textbf{tree-structure normalization}, we use a fully deterministic configuration 
(temperature = 0.0, top-$p$ = 0.7, max tokens = 8192), ensuring stable, reproducible JSON 
reconstruction aligned with the expected schema. 
For \textbf{speaker-profile generation}, we adopt a more expressive setting 
(temperature = 1.2, top-$p$ = 0.9, max tokens = 2048) to allow stylistic variability 
when synthesizing biographical descriptions. 
For \textbf{message refinement}, we employ a moderately stochastic configuration 
(temperature = 0.7, top-$p$ = 0.9, max tokens = 512), balancing stylistic flexibility 
with semantic faithfulness to the draft. 
All calls use the same model (\texttt{Llama 4 Maverick}) 
and a fixed seed (42). 
Complete templates and parameter settings are available in the project repository 
(see the \href{https://github.com/LeonardoBottonaUniTn/demo_conv_creation/blob/eacl_demo/conv_creator/backend/scripts/llm_calls.py}{\texttt{llm\_calls.py}} module). The platform allows users to register API keys for multiple providers (currently Groq and OpenAI) and to switch among models; keys are stored locally and used only for the user's own annotation session.

\subsection{Labelling interface}\label{app:annot_interf}
\begin{figure*}
    \centering
    \includegraphics[width=0.98\linewidth]{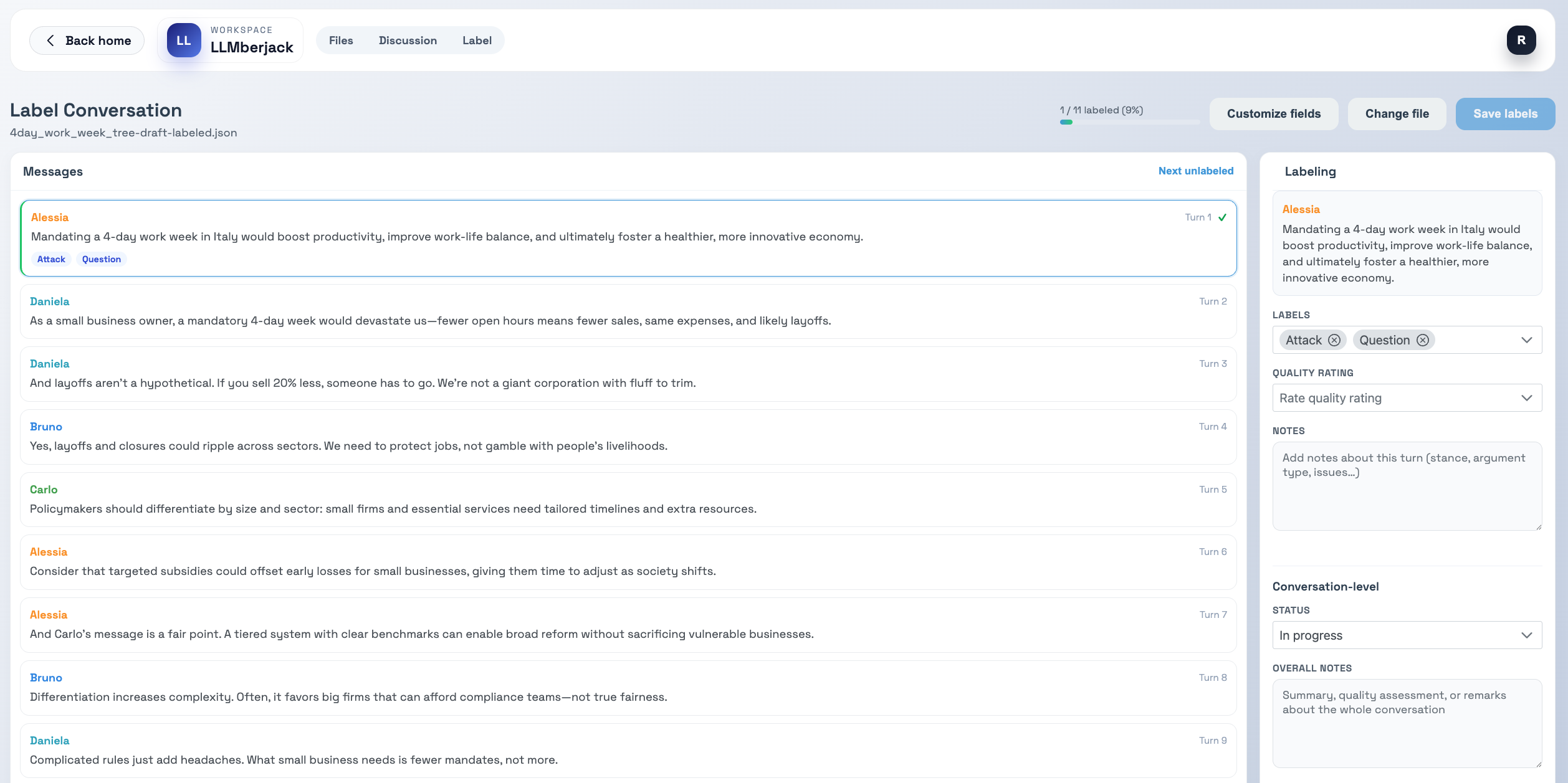}
    \caption{Screenshot of the labelling interface. The left panel shows the conversation, where users can select a message and assign it the labels from the supported label types. The right panel reports the labels assigned at both the message level (for the selected message) and the conversation level.}
    \label{fig:screen_label}
\end{figure*}

In addition to tree visualization and LLM-assisted refinement, \textsc{LLMberjack} provides a labelling window for evaluating linearized conversations, operating at either the message or the conversation level. Annotators can assign free-text string labels, binary (0/1) or Likert-scale ratings. To meet the needs of different labelling tasks, we give the possibility to freely configure the labelling templates. For the future, we plan to include pairwise comparisons between conversations, selecting a preferred candidate or indicating a tie. This is intended to support a complete pipeline of MPC synthetic generation through \textsc{LLMberjack}. A screenshot of the labelling interface is reported in Figure \ref{fig:screen_label}.

\section{Evaluation details}

\subsection{Synthetic Reply Trees} \label{app:reply_trees}

The selected topics for the synthetic reply trees are:

\begin{itemize}
\item[\textbf{T\textsubscript{1}:}] Legalization of marijuana in Italy
\item[\textbf{T\textsubscript{2}:}] Legalization of euthanasia in Italy
\item[\textbf{T\textsubscript{3}:}] Introduction of a four-day work week
\item[\textbf{T\textsubscript{4}:}] Serie A clubs should promote more Italian players rather than foreign stars
\end{itemize}

Since the annotators were Italian, these topics were chosen to reflect debates that are salient within the Italian sociopolitical context. Additionally, we generated a fifth synthetic reply tree on the topic \emph{``Coca-Cola is better than Fanta''}. This tree, along with the MPCs derived from it, was used as tutorial material to familiarize annotators with the platform and the tasks, thereby minimizing platform-related issues during the actual annotation process.

\subsection{Step 1 details} \label{app:step1}

The assignment of tree–visualization pairs was counterbalanced across annotators so that all possible combinations were covered. This design reduces potential topic effects during evaluation and helps identify topics that may be inherently more challenging, while also minimizing annotator-specific variance in the quality assessment.

\subsection{Step 2 guidelines} \label{app:step2}

The assignment of LLM-assisted versus non-assisted refinement was carefully counterbalanced: two couple of annotators (forming one pair) never used the LLM on the same conversation, while the other four possible annotator pairs shared the same setting in exactly half of the cases. This design allows us to evaluate the effect of LLM assistance while controlling for annotator-specific effects and overlapping refinements.

Each annotator refined two conversations for each given topic in a fixed order: first \emph{without} LLM assistance, and then \emph{with} LLM assistance followed by minimal human adjustments. This ordering was chosen to avoid potential bias introduced by prior exposure to LLM-refined content.

The platform has been designed to limit the annotators' freedom on three dimension, with $5$ options each:
\begin{enumerate}
    \item \textbf{Length:} much shorter, slightly shorter, same length, slightly longer, much longer;
    
    \item \textbf{Style:} sarcastic, aggressive, exuberant, cynic, detached;
    
    \item \textbf{Temperament}: neutral, informal, expressive, concise, formal.
    
\end{enumerate}

We asked the annotators to modify the message of only one precise speaker for each topic, so the same speaker for both MPCs. Respectively we asked to make messages more: \textsc{(i.)} \emph{aggressive}, \emph{informal} and \emph{much longer} for T\textsubscript{1}; \textsc{(ii.)} \emph{exuberant}, \emph{expressive} and \emph{same length} for T\textsubscript{2}; \textsc{(iii.)} \emph{cynical}, \emph{concise} and \emph{slightly shorter} for T\textsubscript{3}; \textsc{(iv.)} \emph{detached}, \emph{formal} and \emph{slightly longer} for T\textsubscript{4}. The combination \emph{sarcastic}, \emph{neutral} and \emph{much shorter} was used as ``tutorial'' setting.

\end{document}